\documentclass[10pt,twocolumn]{article}
\usepackage{amssymb}

\usepackage{amsfonts}
\usepackage{amsmath}
\usepackage{tikz}
\usepackage[percent]{overpic}

\DeclareMathOperator*{\tr}{tr}

\newcommand{\Tr}{^\mathrm{T}}

\newcommand{\Xx}{\mathcal{X}}
\newcommand{\Yy}{\mathcal{Y}}

\newcommand{\EE}{\mathbb{E}}
\newcommand{\PP}{\mathbb{P}}

\newcommand{\bb}[1]{\pmb{\mathrm{#1}}}

\title{Bayesian Inference of Bijective Non-Rigid Shape Correspondence}
\author{Matthias Vestner$^{1}$, Roee Litman$^{2}$, Alex Bronstein$^{2,3,5}$, Emanuele Rodol\`{a}$^{1,4}$ and Daniel Cremers$^{1}$ \\
         $^1$Technische Universit\"at M\"unchen, Germany\\
         $^2$Tel-Aviv Univeristy, Israel\\
         $^3$Technion, Israel Institute of Technology, Israel\\
         $^4$Universit\`{a} della Svizzera italiana, Switzerland\\
         $^5$Perceptual Computing Group, Intel, Israel
        }
         
\begin{document}

% \teaser{

%  \centering
%   \caption{Qualitative comparison of methods for pointwise correspondence recovery from a functional map. Current methods such as Nearest Neighbors (NN) and coherent point drift (CPD) suffer from bad accuracy and lack of surjectivity. Applying the proposed Bayesian estimation to either of them gives a guaranteed bijective matching with high accuracy. Left: We visualize the accuracy of the methods by transferring texture from the source shape $\Xx$ to the target shape $\Yy$. Neither Nearest Neighbors nor CPD produce bijective mappings. The lack of surjectivity is visualized by assigning a fixed color (green) to $y\notin im(\Xx)$. Right: The geodesic error (distance between groundtruth and recovered match, relative to the shape diameter) induced by the matching is visualized on the target shape $\Yy$.}
% \label{fig:teaser}
% }

\maketitle

\begin{figure*}
\centering
\begin{overpic}[width=\linewidth]{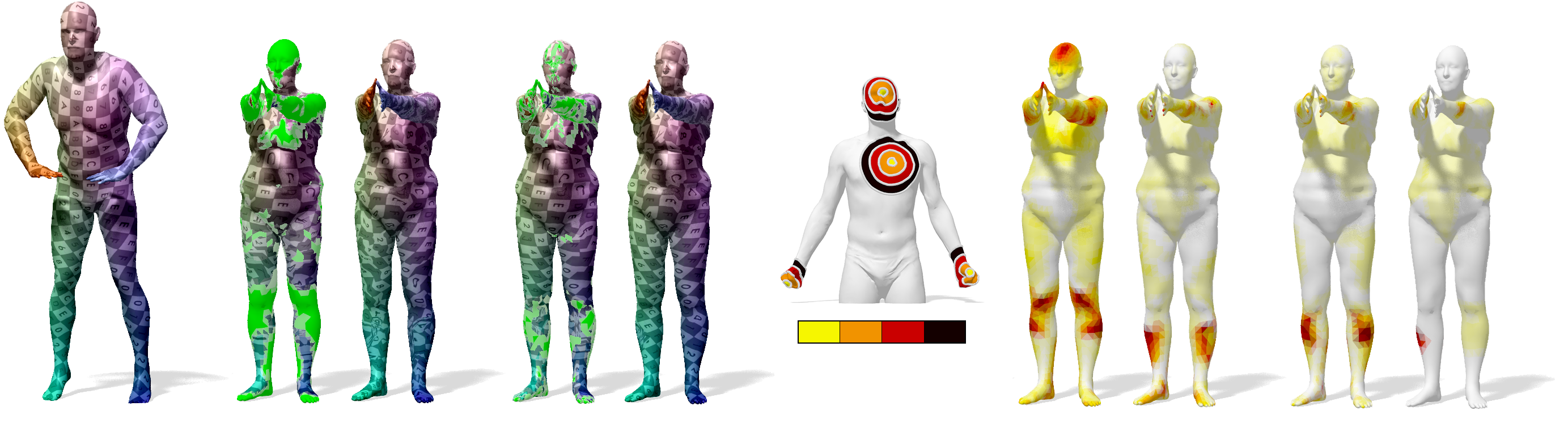}
\put(51.5,3.5){\tiny $1\%$}
\put(54.5,3.5){\tiny $3\%$}
\put(57.5,3.5){\tiny $5\%$}
\put(60.5,3.5){\tiny $7\%$}
\put(54.5,2.25){\tiny $\times \mathrm{diam}$}
\put(17,25.5){\small NN}
\put(21.5,25.5){\small NN+Bayes}
\put(34,25.5){\small CPD}
\put(39,25.5){\small CPD+Bayes}
\put(67,25.5){\small NN}
\put(71,25.5){\small NN+Bayes}
\put(83.5,25.5){\small CPD}
\put(88.5,25.5){\small CPD+Bayes}
\end{overpic}
\caption{Qualitative comparison of methods for pointwise correspondence recovery from a functional map. Current methods such as Nearest Neighbors (NN) and coherent point drift (CPD) suffer from bad accuracy and lack of surjectivity. Applying the proposed Bayesian estimation to either of them gives a guaranteed bijective matching with high accuracy. Left: We visualize the accuracy of the methods by transferring texture from the source shape $\Xx$ to the target shape $\Yy$. Neither Nearest Neighbors nor CPD produce bijective mappings. The lack of surjectivity is visualized by assigning a fixed color (green) to $y\notin im(\Xx)$. Right: The geodesic error (distance between groundtruth and recovered match, relative to the shape diameter) induced by the matching is visualized on the target shape $\Yy$.}
\label{fig:teaser}
\end{figure*}
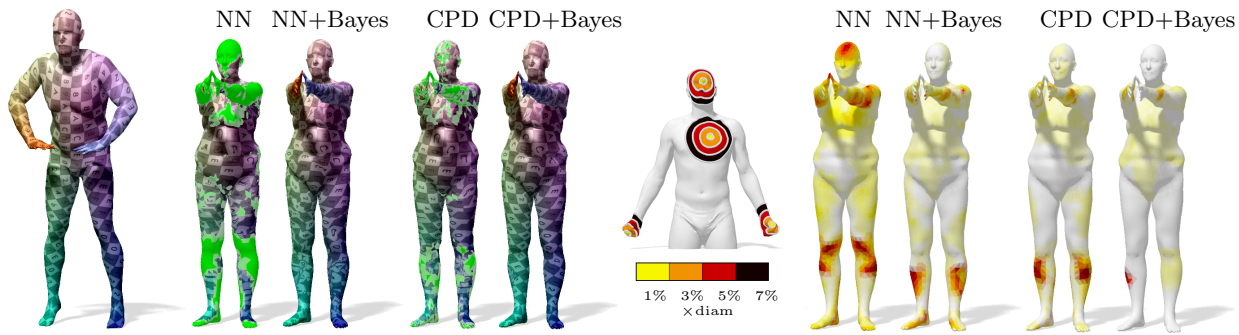

 \begin{abstract}
Many algorithms for the computation of correspondences between deformable shapes rely on some variant of nearest neighbor matching in a descriptor space. Such are, for example, various point-wise correspondence recovery algorithms used as a post-processing stage in the functional correspondence framework. In this paper, we show that such frequently used techniques in practice suffer from lack of accuracy and result in poor surjectivity. We propose an alternative recovery technique guaranteeing a bijective correspondence and producing significantly higher accuracy. We derive the proposed method from a statistical framework of Bayesian inference and demonstrate its performance on several challenging deformable 3D shape matching.
  \end{abstract}

\section{Introduction}

In geometry processing, computer graphics, and vision, estimating correspondence between 3D shapes affected by different transformations is one of the fundamental problems with a wide spectrum of applications ranging from texture mapping to animation \cite{kaick2010survey}. These problems are becoming increasingly important due to the emergence of affordable 3D sensing technology.
Of particular interest is the setting in which the objects are allowed to deform non-rigidly.

\subsection{Related works} 

A traditional approach to correspondence problems is finding a {\em point-wise} matching between (a subset of) the points on two or more shapes. 
{\em Minimum-distortion methods} establish the matching by minimizing some structure distortion, which can include similarity of local features 
%\cite{OvMe*10,BronsteinK10,aubry-et-al-4dmod11,ZaBo*09,litman2011diffusion},
\cite{OvMe*10,BronsteinK10,WKS,ZaBo*09}, %,litman2011diffusion}, 
geodesic \cite{Memoli:2005,bro:bro:kim:PNAS,koltun} or diffusion distances \cite{lafon:05:LOCAL}, %,bro-ghf},  %\cite{bro-ghf}, 
or a combination thereof \cite{torresani2008feature}. %,wang2011discrete}. 
Windheuser et al. \cite{Windheusericcv11} used the thin shell elastic energy of triangles, while
Zeng et al. \cite{zeng2010dense} used higher-order structures.
Typically, the computational complexity of such methods is high, and there have been several attempts to alleviate the computational complexity using hierarchical %\cite{wang2011discrete,sahillioglu2012} 
\cite{sahillioglu2012} 
or subsampling \cite{tevs2011intrinsic} methods. 
Several approaches formulate the correspondence problem as quadratic assignment and employ different relaxations thereof \cite{umeyama1988eigendecomposition,leordeanu2005spectral,rodola2012game,aflalo2015convex,koltun,lipman15}. 
Algorithms in this category typically produce guaranteed bijective correspondences between a sparse set of points, or a dense correspondence suffering from poor surjectivity.

{\em Embedding methods} try to exploit some assumption on the correspondence (e.g. approximate isometry) in order to parametrize the correspondence problem with a few degrees of freedom. Elad and Kimmel \cite{ela:kim:FLATTEN} used multi-dimensional scaling to embed the geodesic metric of the matched shapes into a low-dimensional Euclidean space, where alignment of the resulting ``canonical forms'' is then performed by simple rigid matching (ICP) \cite{ChenMedioni:91:ICP,bes:mck:SURFACEMATCH}. 
%,Mitra:04:ICP}. 
%
The works of \cite{Mateus08,shtern2013matching} used the eigenfunctions of the Laplace-Beltrami operator as embedding coordinates and performed matching in the eigenspace. 
Lipman et al. \cite{Lipman2011,KimLCF10,kim2011blended} used conformal embeddings into disks and spheres to parametrize correspondences between homeomorphic surfaces as M{\"o}bius transformations. By using locally injective flattenings, \cite{aigerman2014lifted} achieve guaranteed bijective matching.
However, the majority of the matching procedures performed in the embedding space often produces noisy correspondences at fine scales, and suffers from poor surjectivity. 

%M{\"o}bius voting \cite{KimLCF10}
%
%blended intrinsic maps \cite{kim2011blended}
%conformal Wasserstein \cite{Lipman2011}

%In the machine learning community, manifold alignment 

%More recently, there is an emerging interest in {\em soft correspondence} approaches.  
As opposed to point-wise correspondence methods, {\em soft correspondence} approaches assign a point on one shape to more than one point on the other. 
Several methods formulated soft correspondence as a mass-transportation problem \cite{Me11,solomon2012soft}. 
Ovsjanikov \emph{et al.} \cite{ovsjanikov2012functional} introduced the {\em functional correspondence} framework, modeling the correspondence as a linear operator between spaces of functions on two shapes, which has an efficient representation in the Laplacian eigenbases. 
This approach was extended in several follow-up works \cite{pokrass2013sparse,kovnatsky15,SGMDS} .
A point-wise map is typically recovered from a low-rank approximation of the functional correspondence by a matching procedure in the representation basis, which also suffers from poor surjectivity. 

\subsection{Main contributions}

As the main contribution of this paper we see the formulation of the intrinsic map \emph{denoising} problem: Given a set of point-wise correspondences between two shapes coming from any correspondence algorithm (for example, using one of the recovery algorithms outlined in Section \ref{sec:recovery}), we consider them as a noisy realization of a latent bijective correspondence. We estimate this bijection using an intrinsic equivalent of the standard minimum mean squared error (MMSE) or minimum mean absolute error (MMAE) Bayesian estimators. To the best of our knowledge, despite their simplicity, these tools have not been previously used for deformable shape analysis.

We show that the considered family of Bayesian estimators leads to a linear assignment problem (LAP) guaranteeing bijective correspondence between the shapes. Despite the common wisdom, we demonstrate that the problem is efficiently solvable for relatively densely sampled shapes by means of the well-established auction algorithm \cite{bertsekas} and a simple multi-scale approach. 

Finally, we present a significant amount of empirical evidence that the proposed denoising procedure consistently improves the quality of the input correspondence coming from different algorithms.

\section{Pointwise map recovery}
\label{sec:recovery}

We start by briefly overviewing several recent techniques used for the computation of pointwise correspondences between non-rigid shapes. We focus on approaches relying on the functional map formalism merely because these techniques produce state-of-the-art results, emphasizing that the proposed algorithm can accept any point-wise correspondence as the input.

We model shapes as connected two-dimensional Riemannian manifolds $\Xx$ (possibly with boundary) endowed with the standard measure $da$ induced by the volume form. Shape $\Xx$ is equipped with the symmetric Laplace-Beltrami operator $\Delta_{\Xx}$, generalizing the notion of Laplacian to manifolds. The manifold Laplacian yields an eigen-decomposition ${\Delta_{\Xx} \phi_i = \lambda_i \phi_i}$ for ${i\geq 1}$, with eigenvalues $0 = \lambda_1 < \lambda_2 \leq \hdots$ and eigenfunctions $\{\phi_i\}_{i\geq 1}$ forming an orthonormal basis of $L^2(\Xx)$.
%
%We denote the space of square-integrable functions on the manifold $\mathcal{M}$ by $L^2(\mathcal{M}) = \{ f: \mathcal{M} \rightarrow\mathbb{R} ~|~ \int_{\mathcal{M}}f^2d\mu <\infty \}$, and use the standard $L^2(\mathcal{M})$ inner product $\langle f, g\rangle_{\mathcal{M}} = \int_{\mathcal{M}}fg d\mu$.
%
Due to the latter property, any function $f\in L^2(\Xx)$ can be represented via the (manifold) Fourier series expansion
\begin{eqnarray}
\label{eq:fourier}
f(x) &=& \sum_{i\geq 1}  \langle f, \phi_i\rangle_{\Xx} \phi_i(x)\,,
\end{eqnarray}
where we use the standard manifold inner product $\langle f,g\rangle_{\Xx}=\int_{\Xx}fg da$.

Consider two manifolds $\Xx$ and $\Yy$, and let $\pi:\Xx\to\Yy$ be a bijective mapping between them. In \cite{ovsjanikov2012functional} it was proposed to consider an operator $T: L^2(\Xx) \rightarrow L^2(\Yy)$, mapping functions on $\Xx$ to functions on $\Yy$ via the composition $T(f) = f \circ \pi^{-1}$. This simple change in paradigm remarkably allows to identify maps between manifolds as linear operators (named {\em functional maps}) between Hilbert spaces.
Because $T$ is a linear operator, it admits a matrix representation with respect to a choice of bases $\{\phi_i\}_{i\geq 1}$ and $\{\psi_i\}_{i\geq 1}$ on $L^2(\Xx)$ and $L^2(\Yy)$, respectively. Assuming the bases to be orthogonal, the matrix $\pmb{C}$ with the elements $(\pmb{C})_{ij} = \langle T(\phi_i), \psi_j \rangle_{\Yy}$ provides a representation of $T$. In particular, by choosing the delta functions supported on the shape vertices as basis functions, one obtains a permutation $\bb{\Pi}$ as a matrix representation for the functional map.

A more compact way to represent $T$ in matrix form is obtained by taking the Laplacian eigenfunctions $\{\phi_i\}_{i\geq 1}$, $\{\psi_i\}_{i\geq 1}$ of the respective manifolds as the choice for a basis. In case $\pi$ is a (near) isometry, the equality $\psi_i=\pm\phi_i \circ \pi^{-1}$ holds (approximately) for all $i\ge 1$, leading to the matrix representation $\pmb{C}$ being diagonally dominant, i.e., $(\pmb{C})_{ij} = \langle T(\phi_i),\psi_j\rangle_{\Yy} \approx \pm\delta_{ij}$.

With this choice, Ovsjanikov et al. \cite{ovsjanikov2012functional} proposed to truncate the matrix $\pmb{C}$ after the first $k \times k$ coefficients as a low-pass approximation of the functional map (typical values for $k$ are in the range $20-300$).
This is especially convenient for correspondence problems, where one is required to solve for $k^2$.
% as opposed to $n^2$ unknowns.
At the same time, in analogy to classical Fourier analysis, the truncation has a {\em blurring} effect on the correspondence. As a result, recovering the original bijection $\pi$ from the spectral coefficients $\pmb{C}$ leads to a non-trivial inverse problem.

Assume shapes $\Xx$ and $\Yy$ have $n$ points each, and let the matrices $\bb{\Phi}, \bb{\Psi} \in \mathbb{R}^{n \times k}$ contain the first $k\ll n$ eigenvectors of the respective Laplacians.
For the sake of simplicity we assume  $\bb{\Phi}$ and $\bb{\Psi}$ to be area-weighted, allowing us to consider the standard dot product in all equations. The expression for $\pmb{C} \in \mathbb{R}^{k\times k}$ can now be compactly written as
\begin{equation}\label{eq:ppp}
\pmb{C} = \bb{\Psi}\Tr \bb{\Pi} \bb{\Phi}\,.
\end{equation}
Note that the matrix $\pmb{C}$ is now a rank-$k$ approximation of $T$. The pointwise map recovery problem \cite{rodola-vmv15}, which is highly underdetermined, consists in finding a $n \times n$ permutation $\bb{\Pi}$ satisfying \eqref{eq:ppp}. The following techniques have been proposed for this purpose.

\paragraph*{Linear assignment problem (LAP).}
If we assume $k=n$, the terms on either side of  \eqref{eq:ppp} have the same rank and the relation can be straightforwardly inverted to yield $\bb{\Pi} = \bb{\Psi} \pmb{C}\bb{\Phi}\Tr$. Since in the truncated setting we have $k \ll n$, the best possible solution in the $\ell^2$ sense can be obtained by looking for a permutation $\bb{\Pi}$ minimizing $- \langle \bb{\Pi} , \bb{\Psi} \pmb{C}\bb{\Phi}\Tr  \rangle_F$. This leads to the equivalent linear assignment problem:
\begin{align}\label{eq:lap}
\min_{\bb{\Pi} \in \{0,1\}^{n \times n}}~& \|\pmb{C}\bb{\Phi}\Tr - \bb{\Psi}\Tr \bb{\Pi}\|_ \mathrm{F}^2\\ \label{eq:doublyStochastic2}
\mathrm{s.t.}~~&\bb{\Pi}\Tr\pmb{1}=\pmb{1}\,,~\bb{\Pi}\pmb{1}=\pmb{1}\,.
\end{align}
The equality between the two expressions comes from the observation that $\|\pmb{C}\bb{\Phi}\Tr - \bb{\Psi}\Tr \bb{\Pi}\|_ \mathrm{F}^2 = \|\pmb{C}\bb{\Phi}\Tr\|_ \mathrm{F}^2 + \|\bb{\Psi}\Tr\|_ \mathrm{F}^2 - 2 \langle \bb{\Psi} \pmb{C}\bb{\Phi}\Tr , \bb{\Pi} \rangle$ for permutation matrices $\bb{\Pi}$. Minimizing with respect to $\bb{\Pi}$, and recalling that $\bb{\Psi}\Tr\bb{\Psi}=\pmb{I}$, yields the equivalence.

The problem above admits an intuitive interpretation. Denoting by $\pmb{e}_i$ the indicator vector having the value 1 in the $i$th position and 0 otherwise, we see that each column of $\bb{\Phi}\Tr$ contains the spectral coefficients $\bb{\Phi}\Tr\pmb{e}_i$ of delta functions $\delta_{x_i}:\Xx\to\{0,1\}$ for $x_i\in\Xx$ and $i=1,\dots ,n$. Hence, the image via $T$ of all indicator functions on $\Xx$ is given by the columns of $\pmb{C}\bb{\Phi}\Tr$. Problem \eqref{eq:lap} seeks for a permutation $\bb{\Pi}$ minimizing the distance between columns of ${\pmb{C}\bb{\Phi}\Tr}$ and columns of $\bb{\Psi}\Tr\bb{\Pi}$ in a $\ell^2$ sense.
%

%Although the problem above admits a polynomial time solution~\cite{kuhn55}, typical values for $n$ (in the order of thousands) make computing this solution prohibitively expensive in practice.

\paragraph*{Nearest neighbors.}
In \cite{ovsjanikov2012functional} the authors proposed to recover a pointwise correspondence between $\Xx$ and $\Yy$ by solving the nearest-neighbor problem
\begin{align}\label{eq:nn1}
\min_{\pmb{P} \in \{0,1\}^{n \times n}}~& \|\pmb{C}\bb{\Phi}\Tr - \bb{\Psi}\Tr \pmb{P}\|_ \mathrm{F}^2\\\label{eq:cl}
\mathrm{s.t.}~&\pmb{P}\Tr\pmb{1}=\pmb{1}\,.
\end{align}
This can be seen as a simplified version of the LAP where the bi-stochasticity constraints \eqref{eq:doublyStochastic2} are relaxed, including all (binary) column-stochastic matrices $\pmb{P}$ in the feasible set. A global solution to \eqref{eq:nn1} can be obtained in an efficient manner by solving for each column of $\pmb{P}$ separately: It is sufficient to seek for the nearest column of $\pmb{C}\bb{\Phi}\Tr$ with respect to each column of $\bb{\Psi}\Tr$.

An immediate consequence of this separable approach is that its minimizers are not guaranteed to be bijections. A balanced version of \eqref{eq:nn1}, obtained by exchanging the roles of $\pmb{C}\bb{\Phi}\Tr$ and $\bb{\Psi}\Tr$ in an alternating fashion was proposed in \cite{rodola-vmv15}, with moderate increase in accuracy.

\paragraph*{Iterative closest point (ICP).}
In \cite{ovsjanikov2012functional} it was additionally proposed to solve for the (not necessarily bijective) $\pmb{P}$ according to the nearest-neighbor approach \eqref{eq:nn1}, followed by a refinement of $\pmb{C}$ via the orthogonal Procrustes problem:
\begin{align}\label{eq:icp}
\min_{\pmb{C}\in\mathbb{R}^{k\times k}}~& \|\pmb{C}\bb{\Phi}\Tr - \bb{\Psi}\Tr \pmb{P}\|_ \mathrm{F}^2\\\label{eq:ortho}
\mathrm{s.t.}~ &\pmb{C}\Tr\pmb{C}=\pmb{I}\,.
\end{align}
The $\pmb{P}$- and $\pmb{C}$- steps are alternated until convergence. In analogy to classical Iterative Closest Point (ICP) refinement \cite{ChenMedioni:91:ICP,bes:mck:SURFACEMATCH} operating in $\mathbb{R}^3$,  this can be seen as a rigid alignment between point sets (columns of) $\bb{\Phi}\Tr$ and $\bb{\Psi}\Tr \pmb{P}$ in $\mathbb{R}^k$.

\paragraph*{Coherent point drift (CPD).}
The orthogonal refinement of \eqref{eq:nn1}, \eqref{eq:icp} assumes the underlying map to be area-preserving \cite{ovsjanikov2012functional}, and is therefore bound to fail in case the two shapes are non-isometric.
Rodol\`{a} et al. \cite{rodola-vmv15} proposed to consider the non-rigid counterpart, for a given $\pmb{C}$:
\begin{align}
\label{eq:proposedEnergyTotal}
\min_{\pmb{P}\in [0,1]^{n\times n}}~& D_\mathrm{KL}(\pmb{C}\bb{\Phi}\Tr,\bb{\Psi}\Tr\pmb{P})
 + \lambda \| \pmb \Omega (\pmb{C}\bb{\Phi}\Tr - \bb{\Psi}\Tr\pmb{P}) \|^2\\
\mathrm{s.t.}~ &\pmb{P}\Tr\pmb{1}=\pmb{1}\,,
\end{align}
where $D_\mathrm{KL}$ denotes the Kullback-Leibler divergence between probability distributions, $\pmb\Omega$ is a low-pass operator promoting smooth velocity vectors, and $\lambda>0$ controls the regularity of the assignment. Problem \eqref{eq:proposedEnergyTotal}  can be seen as  a Tikhonov regularization of the displacement field relating the two sets of spectral coefficients, with a measure of proximity given by the KL divergence between the two. The problem is then solved via expectation-maximization by the coherent point drift algorithm \cite{myronenko10}.

\section{Bayesian map estimation}

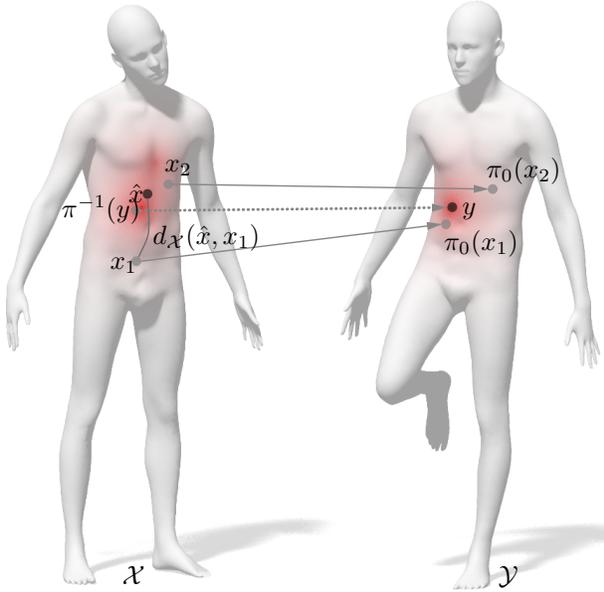
\begin{figure}
\begin{center}
\begin{overpic}[width=1\linewidth]{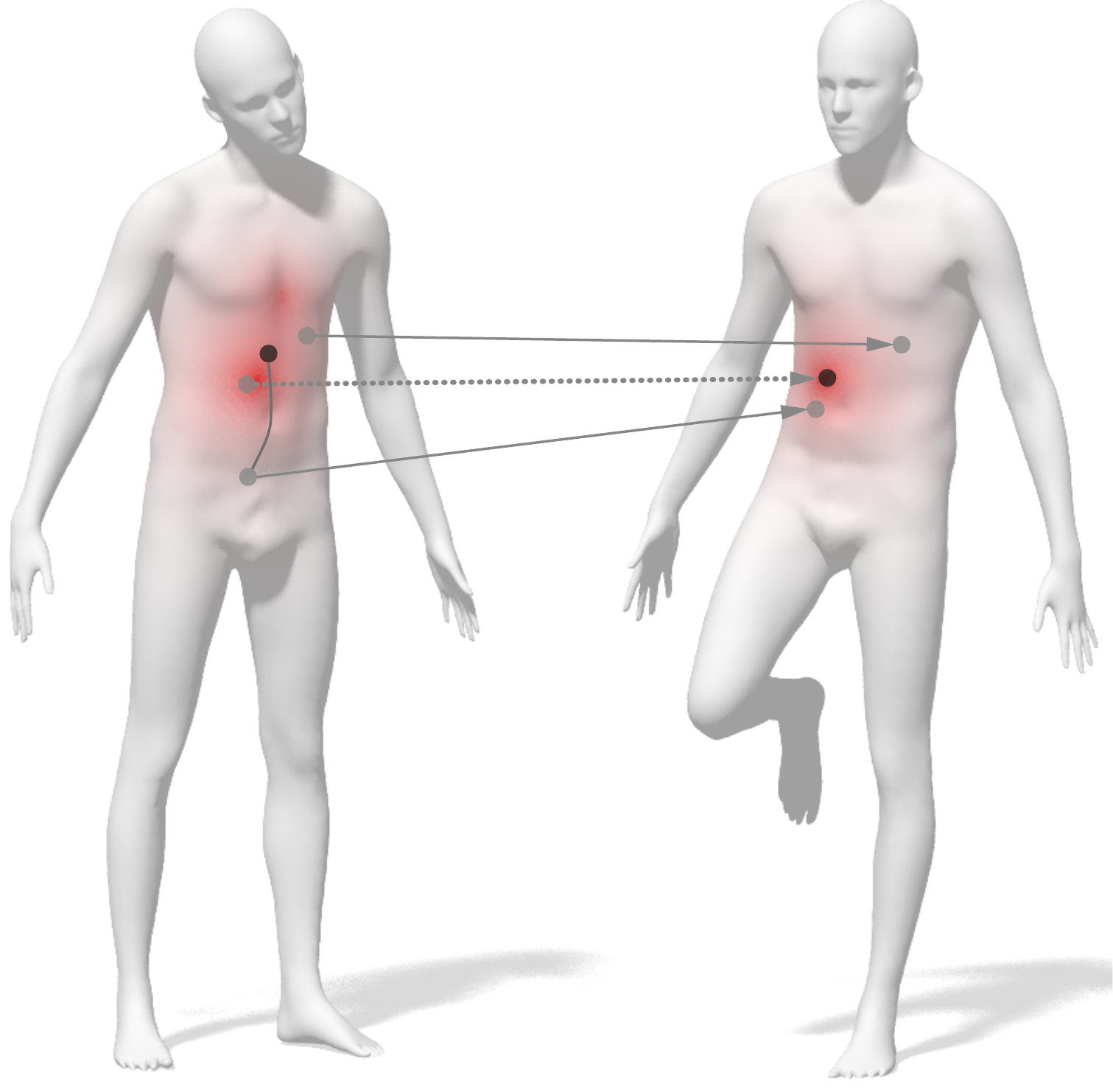}
\put(10,62.5){$\pi^{-1}(y)$}
\put(20,2){$\Xx$}
\put(82,2){$\Yy$}
\put(76,63){$y$}
\put(73,57){$\pi_0(x_1)$}
\put(80,69){$\pi_0(x_2)$}
\put(18,54){$x_1$}
\put(21,65){$\hat{x}$}
\put(27,70){$x_2$}
\put(25,58){$d_\Xx(\hat{x},x_1)$}
\end{overpic}
\end{center}
\caption{Conceptual illustration of the proposed Bayesian estimator. For a fixed point $y$, a Gaussian probability measure on $\Yy$ is pulled back to a measure on $\Xx$ by the given correspondence $\pi_0$. The estimate $\hat{x}$ of the latent preimage $\pi^{-1}(y)$ of $y$ is computed by minimizing the expectation of $d_\Xx^p(\hat{x},\cdot)$ with respect to that measure.}
\label{fig:illustration}
\end{figure}

We describe a Bayesian formulation of bijective map estimation that views the given correspondence as a realization of a random process adding noise to a latent ideal correspondence.
We denote by $\pi : \Xx \rightarrow \Yy$ the latent bijective correspondence between the shapes.
Let $X$ denote a random point on $\Xx$ drawn from a uniform distribution, in the sense that for every measurable set $A \subset \Xx$,
$\PP( X \in A) \propto \mathrm{area}(A)$.
%$$
%\PP( X \in A) = \int_{A}da(x),
%$$
%where $da$ denotes the standard (uniform) %area measure on $\Xx$.
%
Given $X=x$, we denote by the conditional random variable $Y|X=x$ a point on $\Yy$ with a Gaussian distribution with some variance $\sigma^2$ centered at $\pi(x)$ that accounts for the uncertainty in the map.
The Gaussian distribution is interpreted in the sense that for every measurable set $B \subset \Yy$,
$$
\PP( Y \in B | X=x ) \propto \int_{B} \exp\left(-\frac{d_\Yy^2(y,\pi(x))}{2\sigma^2} \right) da(y).
$$

Using Bayes' theorem, we can express the probability density of the conditional random variable $X|Y$ as
\begin{eqnarray*}
f_{X|Y}(x|y) &=& \frac{f_{Y|X}(y|x) f_X(x) }{f_Y(y)} \propto f_{Y|X}(y|x)  \nonumber\\
&\propto & \exp\left(-\frac{d_\Yy^2(y,\pi(x))}{2\sigma^2} \right).
\end{eqnarray*}
Given some (possibly noisy and not necessarily bijective) correspondence $\pi_0 : \Xx \rightarrow \Yy$, we consider $y = \pi_0(x)$ as a realization of $Y|X=x$ for every $x \in \Xx$. Our goal is to estimate the bijection $\pi$ or its inverse $\pi^{-1}$ from these data.

Let us fix some $y \in \Yy$. A Bayesian estimator of $x = \pi^{-1}(y)$ given the observations $\pi_0$ can be expressed as
\begin{eqnarray*}
\hat{x}(y) &=& \mathrm{arg} \min_{\hat{x}} \EE_{X|Y=y} \, d^p_\Xx( X, \hat{x})  \nonumber\\
&=& \mathrm{arg} \min_{\hat{x}} \int_{\Xx} d^p_\Xx( x, \hat{x}) \exp\left(-\frac{d_\Yy^2(y,\pi_0(x))}{2\sigma^2} \right) da(x).
\end{eqnarray*}
In the Euclidean case, the above Bayesian estimator coincides with the minimum mean absolute error (MMAE) for $p=1$ and the minimum mean squared error (MMSE) for $p=2$; in both cases, it has a closed-form solution as the geometric median and the centroid, respectively.
The more general case discussed here can be thought of as the intrinsic counterpart of the median and the centroid.

We estimate the whole inverse map $\pi^{-1}$ by minimizing
\begin{eqnarray}
\hat{\pi}^{-1}  &=& \mathrm{arg} \min_{ \hat{\pi}^{-1} } \int_{\Xx \times \Yy} d^p_\Xx( x, \hat{\pi}^{-1}(y)) e^{-\frac{d_\Yy^2(y,\pi_0(x))}{2\sigma^2} } da(x) da(y).\nonumber\\
\label{eq:BayesianEst}
\end{eqnarray}
over all bijections $\hat{\pi}^{-1} : \Yy \rightarrow \Xx$. Note that due to the additional constraint that $\hat{\pi}^{-1}$ has to be a bijection, the estimation cannot be done for each point $y$ independently.
We also observe that iterating the process several times consistently improves the estimated map accuracy.

Finally, we note that when $\pi$ is area-preserving or, more generally, scales the metric uniformly, the estimator (\ref{eq:BayesianEst}) can be equivalently rewritten in terms of $\pi$ as
\begin{eqnarray}
\hat{\pi}  &=& \mathrm{arg} \min_{ \hat{\pi} : \Xx \rightarrow \Yy } \int_{\Xx \times \Xx} d^p_\Xx( x, \xi ) e^{-\frac{d_\Yy^2(\hat{\pi}(\xi),\pi_0(x))}{2\sigma^2} } da(x)da(\xi).
\label{eq:BayesianEst1}
\end{eqnarray}
It is worthwhile mentioning that while being natural, the assumption of uniform prior distribution of $X$ on $\Xx$ (embodied in the use of the standard area measure in the above integral) can be replaced by other measures emphasizing regions where errors are less tolerable. Also, non-Gaussian noise models may be more suitable for data coming from a specific correspondence algorithm. We defer these interesting questions to future study.

\paragraph*{Discretization.}
We consider the discretization of (\ref{eq:BayesianEst1}). We assume the shape $\Xx$ to be discretized at $n$ points with the corresponding discrete area elements $a_i$ and pairwise geodesic distance matrix $\bb{D}_\Xx$. Similarly, the shape $\Yy$ is discretized as the same number of points, and its pairwise distance matrix is denoted by $\bb{D}_\Yy$.

The bijective correspondence is represented by the $n \times n$ permutation matrix $\hat{\bb{\Pi}} $ sought by minimizing
\begin{eqnarray}
\hat{\bb{\Pi}} = \arg \min_{\bb{\Pi}} \tr( \bb{\Pi}\Tr \bb{P} \bb{\Gamma} ),
\label{eq:BayesianEst_Discrete}
\end{eqnarray}
where $\bb{P}$ is an $n \times n$ matrix with the elements
$$
(\bb{P})_{ij} = \exp\left(-\frac{(\bb{D}_\Yy)_{\pi_0(i),j}^2}{2\sigma^2} \right),
$$
and $\bb{\Gamma}$ is an $n \times n$ matrix with
$(\bb{\Gamma})_{ij} = (\bb{D}_\Xx)_{ij}^p a_i a_j$.
Note that (\ref{eq:BayesianEst_Discrete}) is a linear assignment problem (LAP). For directly solving the LAP with the specific structure of the score matrix given by $\bb{P}\bb{\Gamma}$, we found the auction algorithm \cite{bertsekas} to perform the best in practice. Its average runtime complexity is $\mathcal{O}(n^2 \log n)$, with $\mathcal{O}(n^2)$ storage complexity if a full score matrix is used. On regular hardware, this translates to several seconds for $n \sim 2.5 \times 10^{3}$, which quickly grows to $20$ seconds for $n=4\times 10^3$ and almost $10$ minutes for $n=12 \times 10^3$, taking tens of gigabytes of memory. We therefore conclude that directly solving the full LAP is practical for $n \lesssim 10^4$, and in the following section propose a multi-scale scheme that can scale to much larger numbers of points.

Another computational bottleneck stems from the computation of pairwise geodesic distances.
For example, using fast marching \cite{kimmel1998computing} the computation requires $\mathcal{O}(n^2 \log n)$ computations and $\mathcal{O}(n^2)$ storage. While the computations can be thoroughly parallelized and executed on a GPU, reducing the complexity by orders of magnitude \cite{weber2008parallel}, the storage of a full distance matrix is still prohibitive for $n \gtrsim 10^4$.
However, since geodesic distance maps are almost everywhere smooth with constant gradient, their approximation in a truncated harmonic basic is optimal in the $\ell_2$ sense \cite{aflalo2013spectral}. Instead of storing an $n \times n$ matrix $\bb{D}_\Xx$, we store the $k \times n$ representation coefficient matrix
$$
\bb{A}_\Xx = {\bb{\Phi}}\Tr  \bb{D}_\Xx,
$$
where ${\bb{\Phi}}$ contains the first $k \ll n$ eigenfunctions of the Laplace-Beltrami operator on $\Xx$.
In order to "decompress" the $i$-th row of $\bb{D}_\Xx$ used in the computation of the LAP score, the corresponding column of $\bb{A}_\Xx$ is multiplied from the left by $\tilde{\bb{\Phi}}$.

It is also worthwhile mentioning that while the geodesic metric is a natural candidate to compute intrinsic distances on a manifold, the proposed estimator can work with other choices. For example, diffusion distances \cite{lafon:05:LOCAL} or other approximations of the geodesic distances \cite{crane2013geodesics} are likely to work equally well while being better amenable both for faster computation and more compact storage.

\paragraph*{Multiscale solution.}
\label{sec:multiscale}
In order to reduce the computation and storage complexity associated with the direct solution of the LAP for large values of $n$, we adopt a multi-scale strategy.
Both shapes are discretized in a hierarchical fashion using farthest point sampling, while the distances and the harmonics are calculated at the finest scale and sub-sampled.

First, a full LAP (\ref{eq:BayesianEst_Discrete}) is solved at a coarse scale. The produced correspondence is interpolated to the next scale and is used as the input correspondence $\pi_0$ to the LAP. While numerous interpolation techniques exist, we found that simple nearest neighbour interpolation produces satisfactory results.
At the finer scale, the space of possible bijections $\hat{\pi}(i)$ is restricted to the points falling into a fixed radius $r$ around each $\pi_0(i)$ (note that $r$ has to be larger than the coarse sampling radius). This is equivalent to assigning infinite score to the prohibited permutations. For a sufficiently small $r$, this strategy results in sparse score matrices, with density significantly lower than $1\%$.

\section{Experiments}

\begin{figure}
\centering
\includegraphics[width=\linewidth]{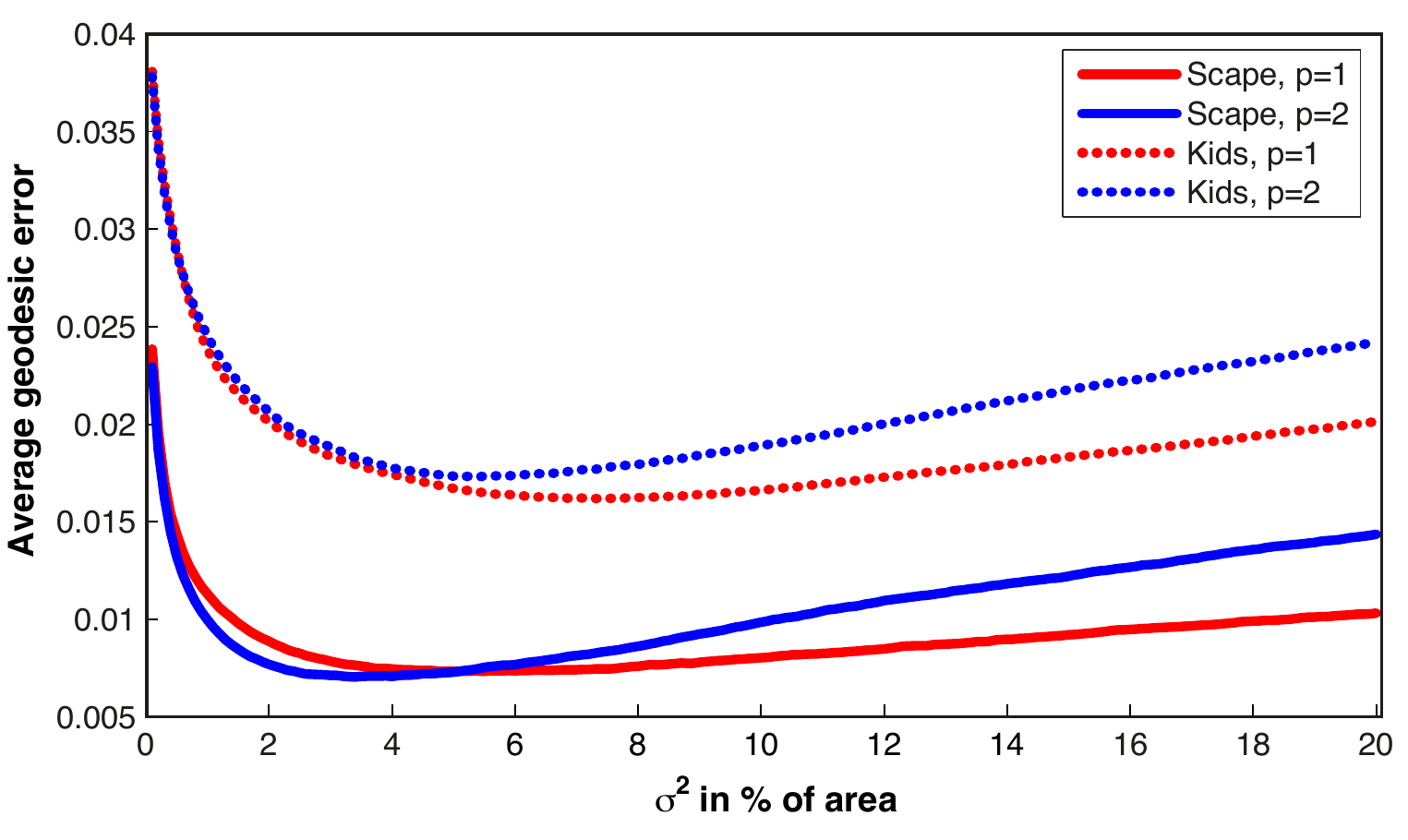}
\caption{Dependency on parameters: We evaluate the influence of the parameters p and $\sigma$ on the quality of the denoised matching. We initialize with a noisy matching coming from a nearest neighbor search in descriptor space. The evaluation is done on two datasets with different global scales, namely KIDS and SCAPE.}
\label{fig:parameter}
\end{figure}

We start by evaluating the influence of the parameters $p\in\{1,2\}$ and $\sigma$ on the quality of the Bayesian estimator \eqref{eq:BayesianEst}. We initialize with noisy correspondences coming from a nearest neighbour \eqref{eq:nn1} result and evaluate on two datasets with different global scales, see Figure \ref{fig:parameter}.  The optimal choice of $\sigma^2$ amounts to approximately 6\% of the target shapes area for both choices of $p$ but the quality is shown to be stable in a vicinity.

We test our method recover bijections from on two types of initialization, namely \textit{functional maps of low rank} and \textit{sparse correspondences}.

We conduct quantitative experiments on the FAUST dataset \cite{faust} (7K vertices) and on downsampled versions of the SCAPE \cite{SCAPE} and KIDS \cite{rodoladense} datasets (1K vertices). As quantitative quality criteria we evaluate the geodesic errors, the run times and the lack of surjectivity of the different methods. We further show that our approach can directly tackle shapes having more then 10K vertices.

\subsection{Recovery from a functional map}

In this set of experiments the low rank approximation is given in terms of a functional map of different ranks in the harmonic basis. Comparisons are done against nearest neighbors (NN) \eqref{eq:nn1}, bijective NN \eqref{eq:lap}, ICP \eqref{eq:icp} and CPD \eqref{eq:proposedEnergyTotal}.

\paragraph*{Approximation of the groundtruth.}

\begin{figure}
\centering
\includegraphics[width=\linewidth]{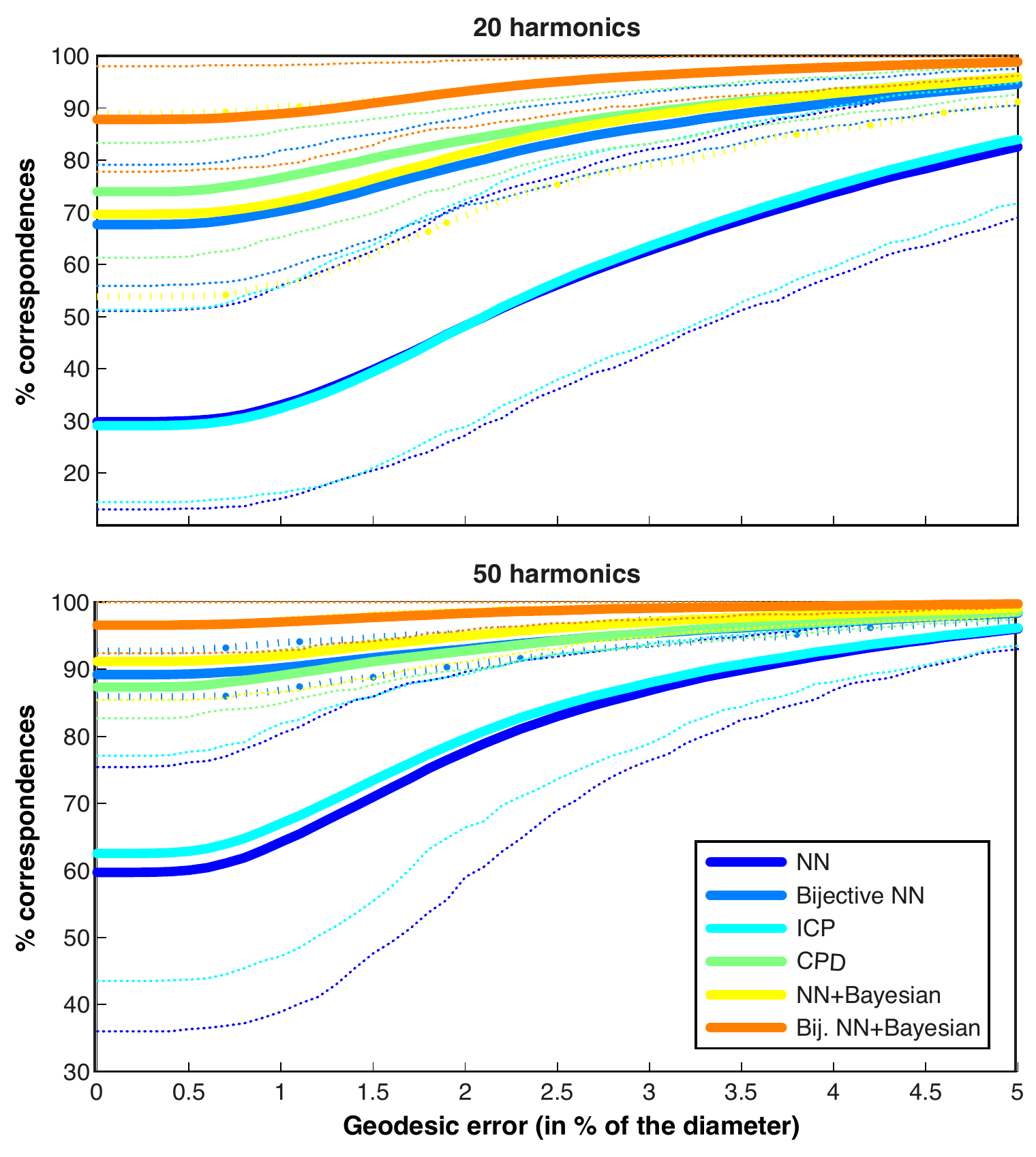}
\caption{Recovering the groundtruth matching from a functional map with rank deficiency. We matched 70 pairs from the near-isometric SCAPE dataset (1K). Plotted are the histograms of geodesic errors (solid line: mean; 90\% of the matched pairs produce results between the dotted lines). All methods boost their quality with increasing numbers of eigenfunctions. Denoising the results of nearest neighbors (yellow) outperforms the state of the art method (green) while having only a fraction of its runtime (Table \ref{tab:run_times}). Even better results are achieved when initializing the Bayesian estimator with the result of bijective NN (orange).}
\label{fig:scape_analysis}
\end{figure}

\begin{figure}
\centering
\includegraphics[width=\linewidth]{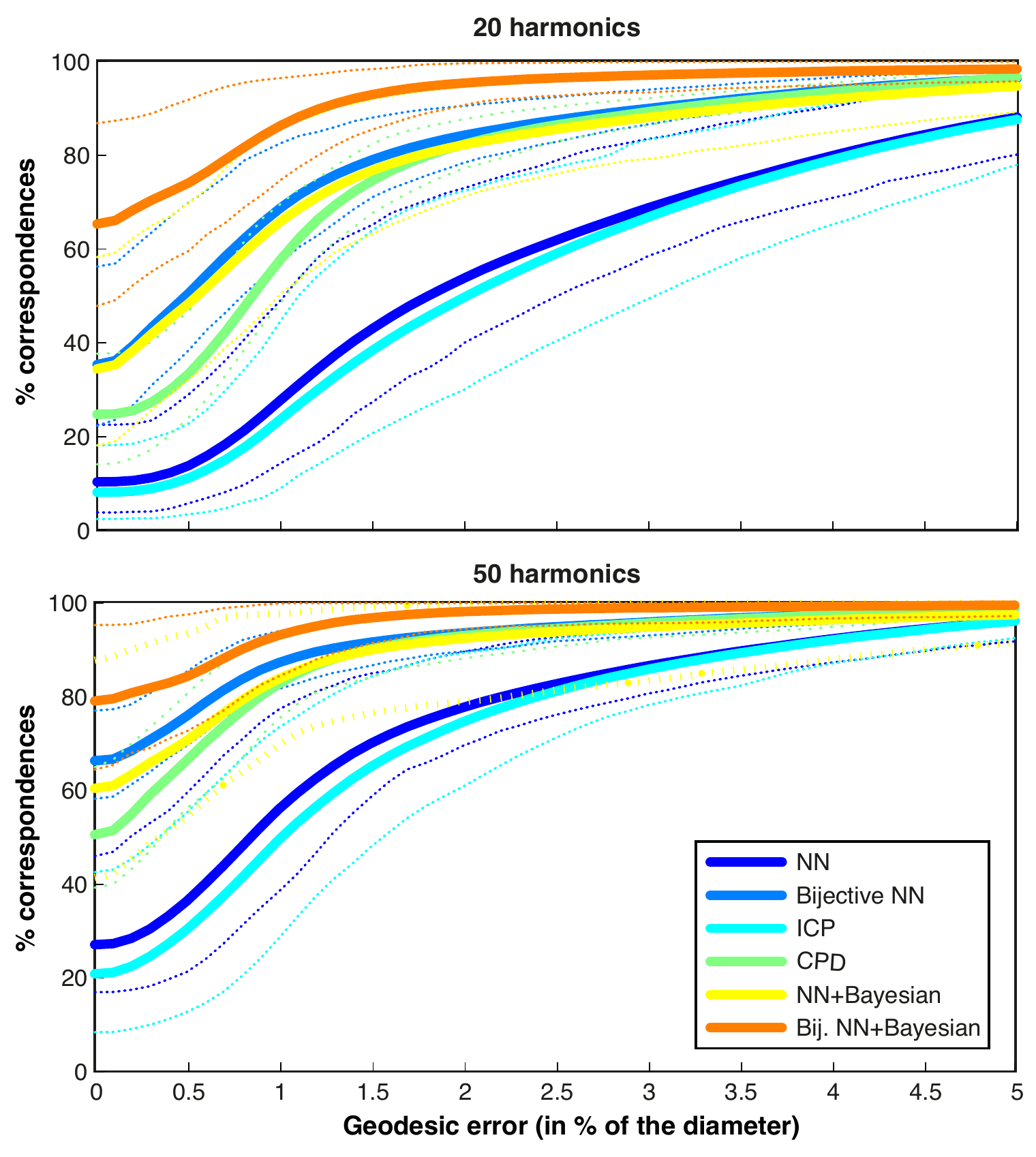}
\caption{Evaluations from Figure \ref{fig:scape_analysis} repeated on the higher resolution FAUST shapes. A single iteration of our denoising algorithm boosts the performance of the simple nearest neighbor approach above the state of the art (green). Notice that in this scenario bijective NN is giving better results than CPD; denoising this matching leads to $80\%$ exact matches averaged over all pairs.}
\label{fig:faust_analysis}
\end{figure}

\begin{figure}
\centering
\includegraphics[width=\linewidth]{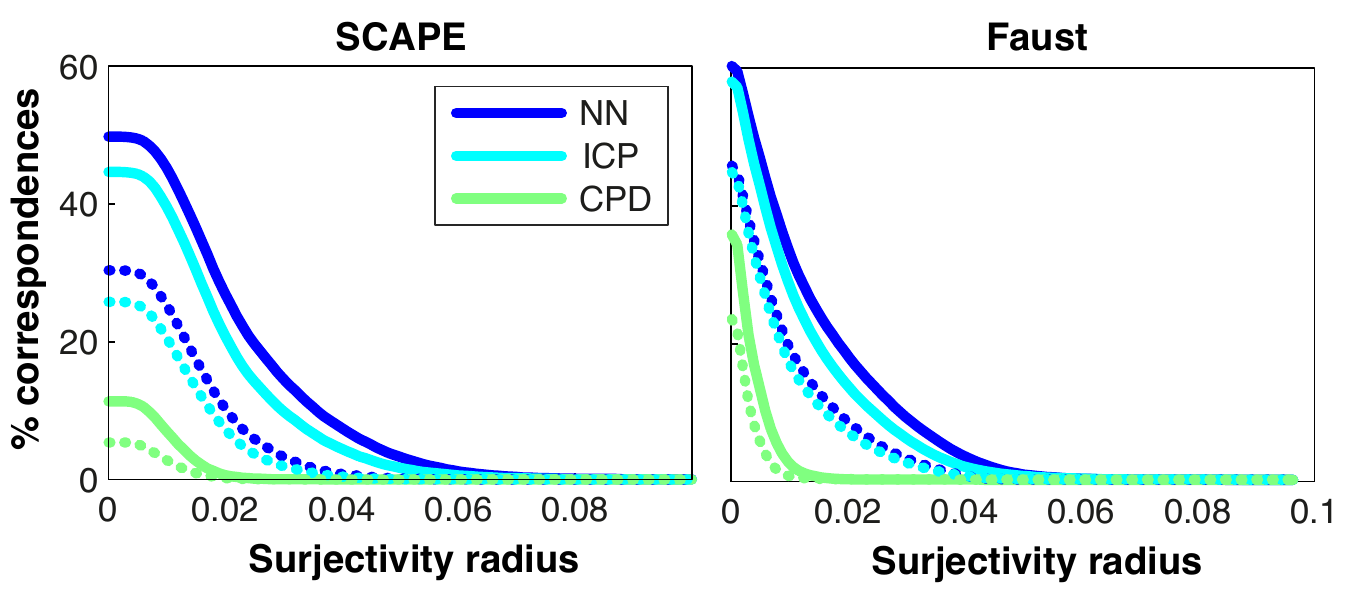}
\caption{Histogram of distances on the target shape to the image of the source shape calculated on SCAPE (left) and FAUST (right) for matching with $20$  (solid) and $50$ (dotted) harmonics. In particular, the intersection with the $y$-axis tells the percentage of points on the target that lack a preimage. For LAP-based approaches imposing bijectivity the image of the source covers the target.}
\label{fig:surjectivity}
\end{figure}

\begin{figure}
\centering
\includegraphics[width=\linewidth]{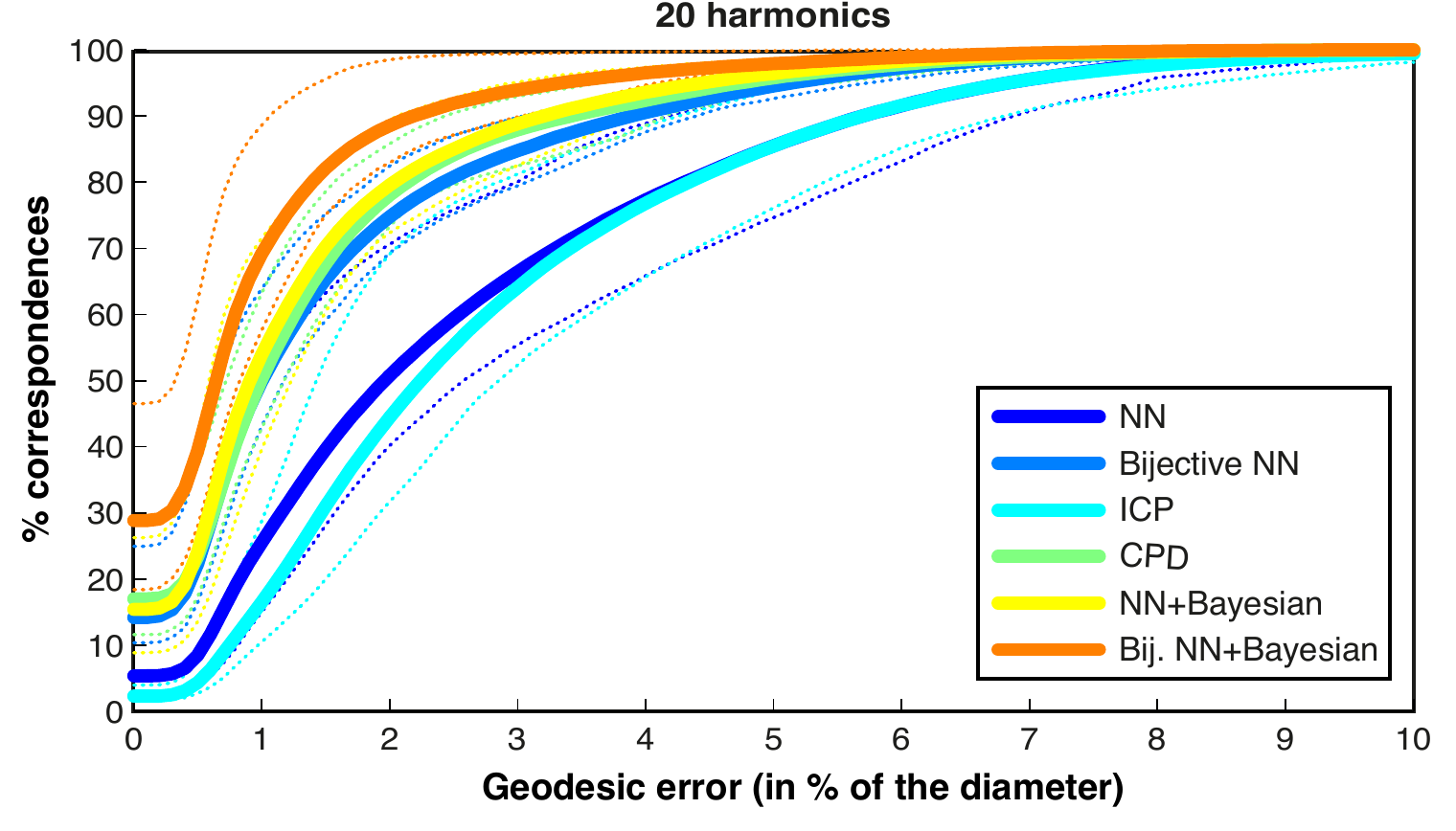}
\caption{Our method can be applied to recover correspondences between high resolution shapes. Although the percentage of exact hits decreases with the increase of the sampling density, the geodesic errors are compelling. Due to the significant increase in runtime  particularly of CPD this experiment was only done on a subset of the pairs from the SCAPE 12.5K benchmark.}
\label{fig:12K}
\end{figure}

Here we construct the low rank functional map using the known groundtruth correspondences between the shapes. This is supposed to be the ideal input for all the competing methods. As the input to our method we use the matchings found by nearest neighbors and its bijective version.
We show quantitative comparisons on 71 pairs from the SCAPE dataset (near isometric, 1K vertices) and 100 pairs from the FAUST dataset (including inter-class pairs, 7K vertices). In Figures  \ref{fig:scape_analysis} and  \ref{fig:faust_analysis} we compare the accuracy, in Figure \ref{fig:surjectivity} the lack of surjectivity is analyzed.
We only show the performance of a single application of the Bayesian estimator yet adumbrate experiments with multiple iterations in the following sections. Even after one iteration, our method outperforms the state of the art method \eqref{eq:proposedEnergyTotal} as well in accuracy as in run time (Table \ref{tab:run_times}). Even on shapes having more then 10K vertices just one iteration of the Bayesian estimator gives very good results, as can be seen in Figure \ref{fig:12K}. Memory consumption and run times however limit the direct applicability of the single-scale Bayesian estimator.% One application of the Bayesian estimator on the centaur shape in Figure \ref{fig:centaur} took more then 3 hours.

\paragraph*{Using a functional map coming from an optimization process.}

\begin{figure}
\centering
\includegraphics[width=\linewidth]{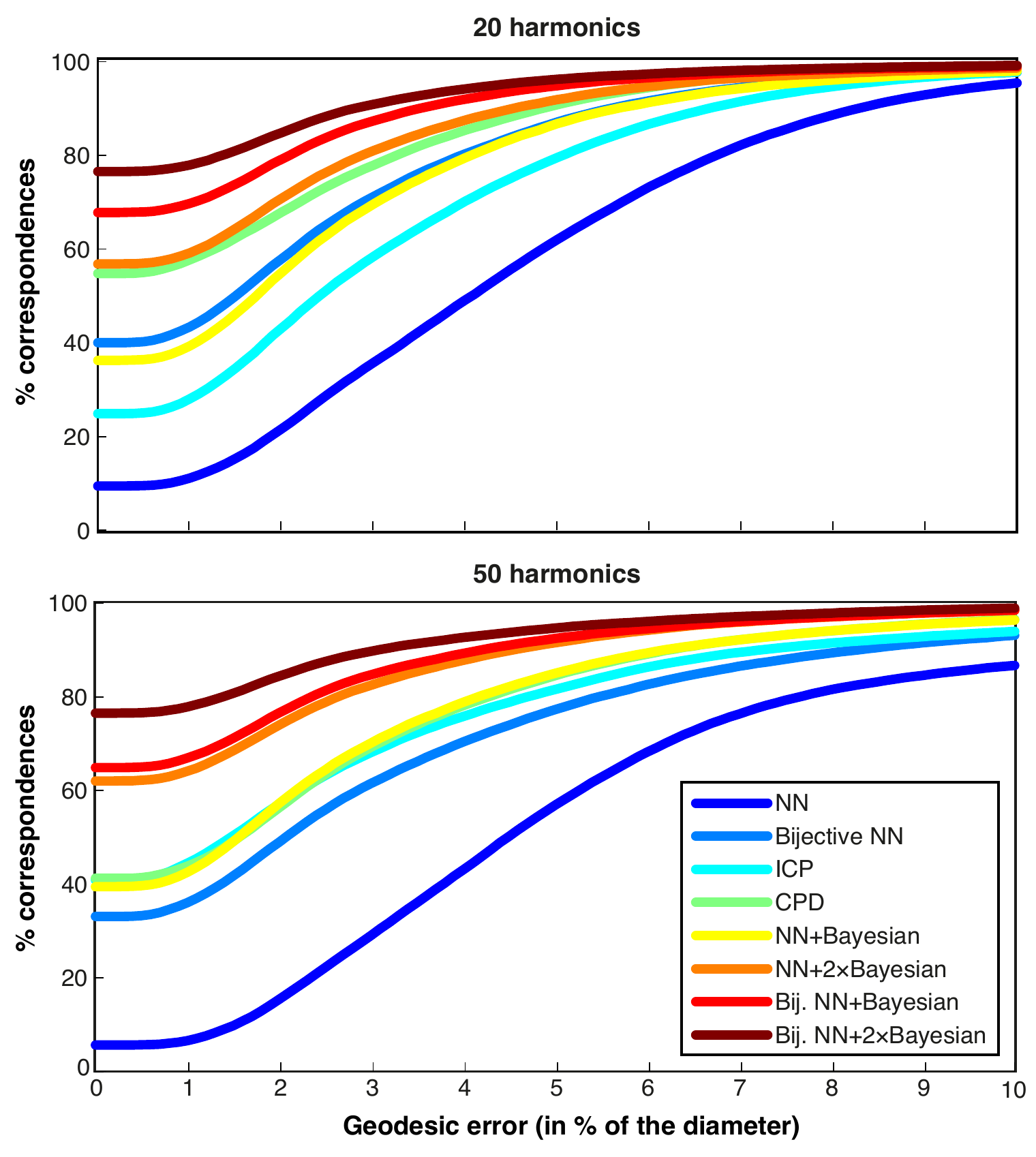}
\caption{Evaluation on realistic input data. We test different recovery methods on a low rank functional map coming from an optimization process. Again our method gives the best results. Iterating the Bayesian estimator improves the performance even more.}
\label{Fig:realC_scape}
\end{figure}

We follow the approach from \cite{pokrass2013sparse} to construct a realistic functional map matching, which typically requires region features as input. These features were detected using the consensus-segmentation method proposed in \cite{rodola2014robust}, and the resulting regions were matched by intersection w.r.t. the ground-truth. Both of these two methods were executed with the same parameters as in their publicly available implementation.
In Figure \ref{Fig:realC_scape} this initialization is evaluated on the SCAPE dataset.

\begin{table}[]
\centering
\begin{tabular}{r|c|c|c|c|}
 $n$& $1000$ & $1000$& $6890$& $6890$  \\
 $k$& $20$ & $50$ & $20$ & $50$\\
 \hline
 Nearest neighbors& 0.04 & 0.06 & 1.35 & 2.88 \\
 Bijective NN&2.79 & 2.30 & 463.66 & 253.03\\
 ICP& 0.14 &0.24  & 12.72 & 30.08\\
 CPD&4.79 & 4.67 & 1745.06 & 2085.65 \\
 NN + Bayesian& 1.75 & 1.28 & 382.86 &244.10\\
 Bij. NN + Bayesian& 4.06 & 3.44 & 746.00 & 440.94
\end{tabular}
\caption{Average runtimes in seconds. We compare the runtimes of different recovery methods. Given the rank $k$ of a functional map approximating the correspondence between shapes sampled at $n$ points each, we report the time it takes to obtain a dense matching. See Figures \ref{fig:scape_analysis}-\ref{Fig:realC_scape} for evaluations of accuracy. Notice that while linear assignment problems are known to be time demanding to solve for larger numbers of variables, the most dramatic increase of run time occurs when applying CPD.}
\label{tab:run_times}
\end{table}

\subsection{Recovery from a sparse correspondence}

\begin{figure}
\centering
\includegraphics[width=\linewidth]{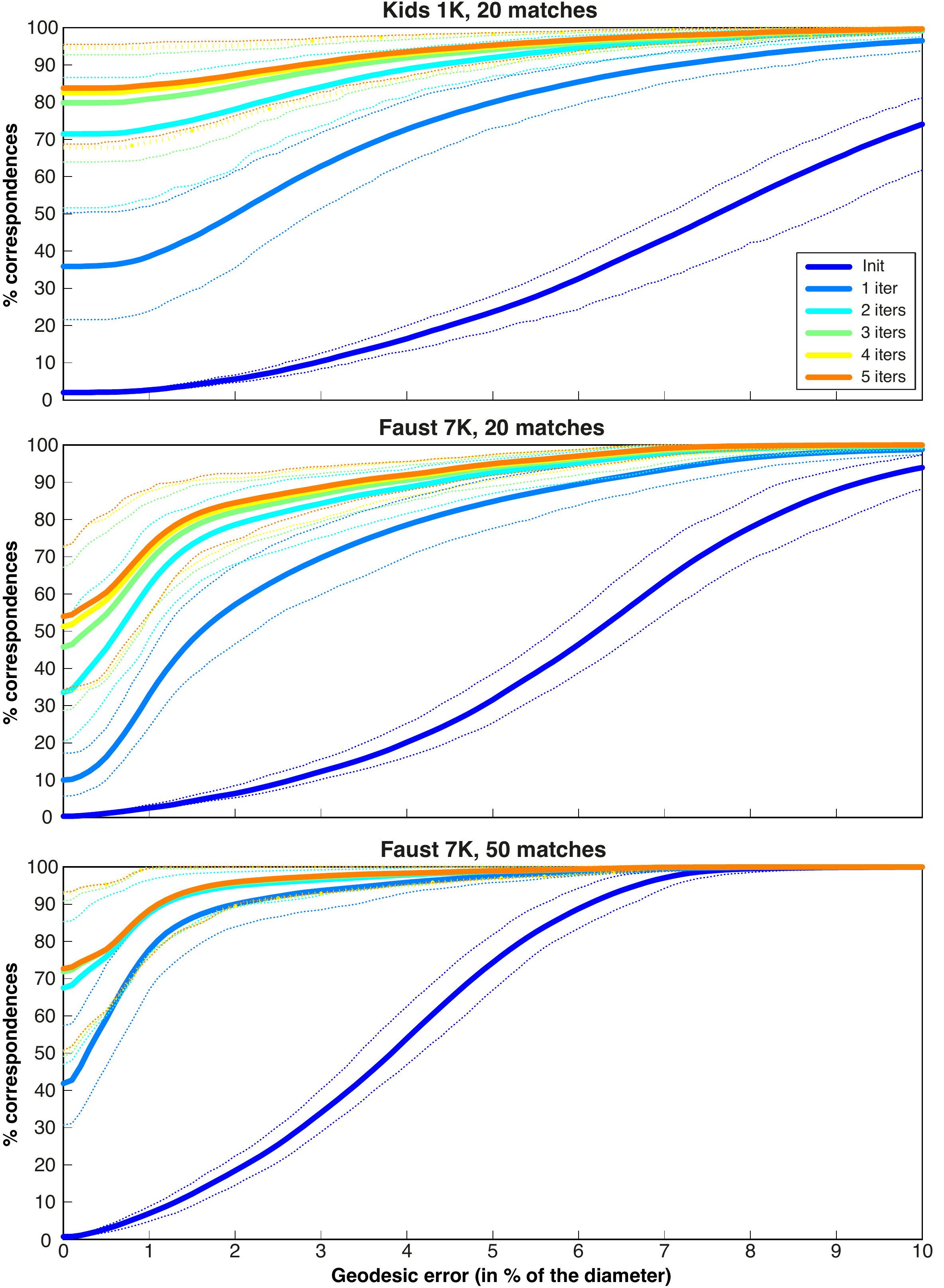}
\caption{Correspondence accuracy on the KIDS 1K (top plot) and FAUST 7K (middle and bottom plots).
Dense matches were produced by nearest-neighbor interpolation from $20$ (top and middle) and $50$ (bottom) sparse matches and used as the initialization of the Bayesian estimator iterated up to five times.}
\label{fig:sparse}
\end{figure}

\begin{figure*}
\centering
\includegraphics[width=0.8\linewidth]{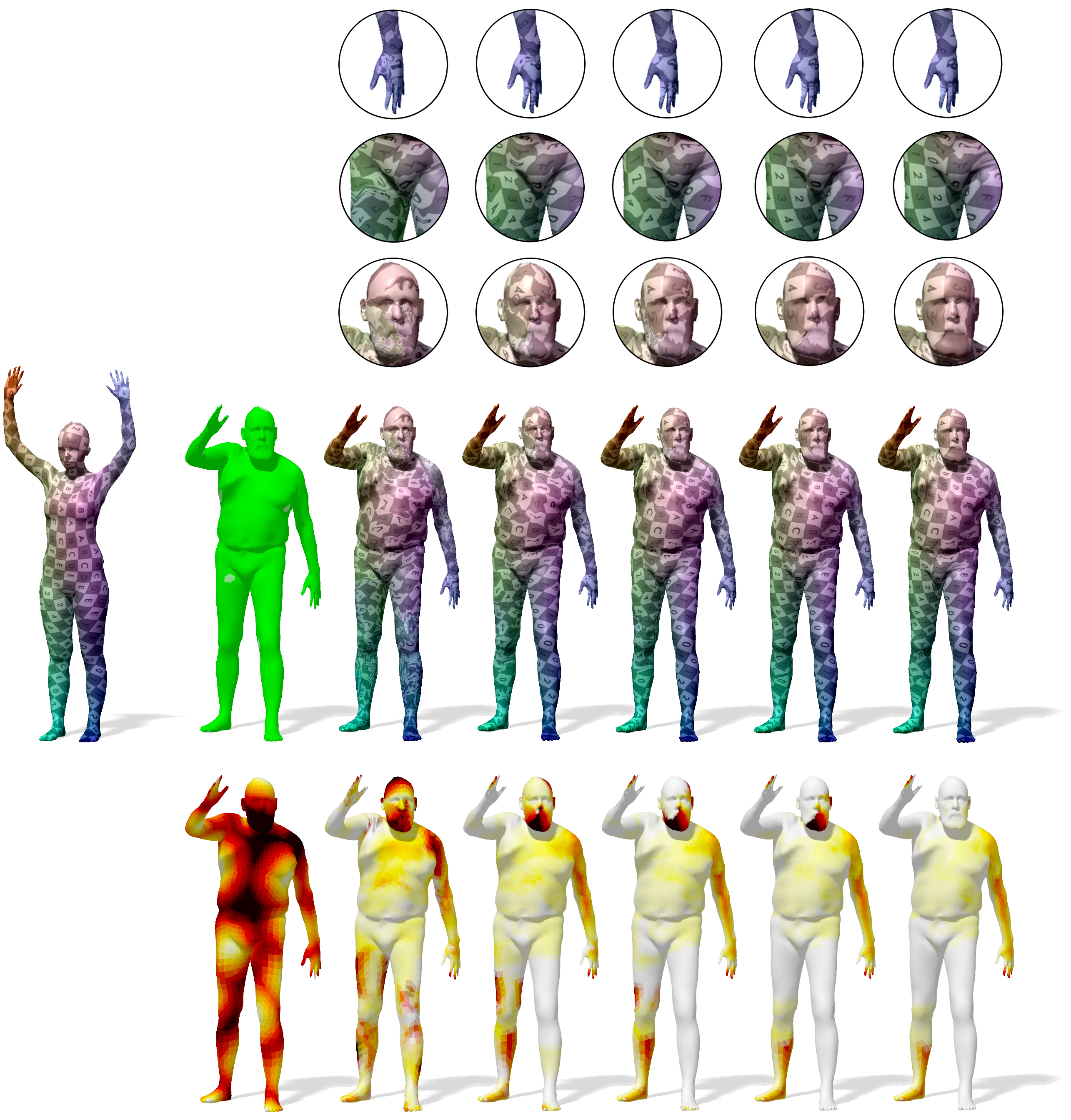}
\caption{Obtaining a dense and bijective matching of high quality by iterative application of the Bayesian estimator on a sparse correspondence. The sparse correspondence is interpolated by assigning each point the on the source shape the same image as its nearest neighbor from the sparse set (second column). This induces large geodesic errors for points being far away from the sparse set (bottom row, left) and is far from being surjective (second row from the bottom, left). Each iteration of the Bayesian estimator (five rightmost columns) yields better results, as illustrated by the magnified fragments.}
\label{fig:iteration_renderings}
\end{figure*}

In this set of experiments the low rank approximation is given in terms of sparse correspondences between high resolution shapes. This type of input can for instance be obtained by minimizing energies under $l^1$-constraints, such as \cite{rodola2012game}, or appears in multiresolution settings.
We make use of groundtruth correspondences of a few points and interpolate the matching with the technique described in the caption of Figure \ref{fig:sparse}. Figure \ref{fig:iteration_renderings} illustrates how iterations of the Bayesian estimator improve the matching.

\section{Conclusion}

We considered the problem of bijective correspondence recovery by means of denoising a given set of matches coming from any of the existing algorithms (including those not guaranteeing bijection, or producing sparse correspondences). Viewing the denosing as a Bayesian estimation problem, we formulated the intrinsic equivalent of the mean and median filters frequently employed in signal processing, with the additional constraint of bijectivity embodied through an LAP.

We find surprising the fact that such a simple idea demonstrates a consistent improvment in the correspondence quality in all experiments we have conducted. We believe that tools from estimation theory that have been heavily used in other domains of science and engineering might be very useful in shape analysis, and invite the community to further explore this direction. 

Of special interest are the choice of the loss function in the posterior expectation (which in this paper was restricted to the absolute and squared distance), the prior distribution of $X$ (which we assumed uniform), and the noise distribution (which was assumed Gaussian). Alternative estimators making use of Bayesian statistics, such as maximum a posteriori (MAP) estimators, should also be explored.

\bibliographystyle{plain}

\bibliography{literature}

\end{document}